\documentclass[conference,anonymous]{IEEEtran}
\IEEEoverridecommandlockouts

\usepackage{cite}
\usepackage{amsmath,amssymb,amsfonts}
\usepackage{algorithmic}
\usepackage{graphicx}
\usepackage{textcomp}
\usepackage[table]{xcolor}
\definecolor{lightred}{RGB}{255, 200, 200}
\definecolor{lightorange}{RGB}{255, 230, 180}
\definecolor{citeblue}{RGB}{87,145,220}
\usepackage[colorlinks=true,citecolor=citeblue]{hyperref}
\usepackage{booktabs}
\usepackage{multirow}

\def\BibTeX{{\rm B\kern-.05em{\sc i\kern-.025em b}\kern-.08em
    T\kern-.1667em\lower.7ex\hbox{E}\kern-.125emX}}
\makeatletter
\g@addto@macro\itemize{\setlength{\itemsep}{1pt}\setlength{\parsep}{0pt}}
\g@addto@macro\enumerate{\setlength{\itemsep}{1pt}\setlength{\parsep}{0pt}}
\makeatother
\begin{document}

\title{
AlignGS: Aligning Geometry and Semantics for Robust Indoor Reconstruction from Sparse Views
}
\author{
Yijie Gao$^{1*}$,
Houqiang Zhong$^{1*}$,
Tianchi Zhu$^{1,3}$,
Zhengxue Cheng$^{1}$,
Qiang Hu$^{2\dagger}$,
Li Song$^{1\dagger}$ \\
$^{1}$ School of Information Science and Electronic Engineering, Shanghai Jiao Tong University, Shanghai, China \\
$^{2}$ Cooperative Mediant Innovation Center, Shanghai Jiao Tong University, Shanghai, China\\
$^{3}$ SJTU Paris Elite Institute of Technology, Shanghai Jiao Tong University, Shanghai, China\\
\{gaoyijie,zhonghouqiang,zhutianchi,zxcheng,qiang.hu,song\_li\}@sjtu.edu.cn
\thanks{This work is supported by National Natural Science Foundation of China (62571322, 62431015, 62271308), STCSM(24ZR1432000, 24511106902, 24511106900, 22DZ2229005), 111 plan(BP0719010), and State Key Laboratory of UHD Video and Audio Production and Presentation. $^*$These authors contribute equally. $^\dagger$Corresponding author.}
}
\maketitle
\begin{abstract}
The demand for semantically rich 3D models of indoor scenes is rapidly growing, driven by  applications in augmented reality, virtual reality, and robotics. However, creating them from sparse views remains a challenge due to geometric ambiguity. Existing methods often treat semantics as a passive feature painted on an already-formed, and potentially flawed, geometry. We posit that for robust sparse-view reconstruction, semantic understanding instead be an active, guiding force. This paper introduces AlignGS, a novel framework that actualizes this vision by pioneering a synergistic, end-to-end optimization of geometry and semantics. Our method distills rich priors from 2D foundation models and uses them to directly regularize the 3D representation through a set of novel semantic-to-geometry guidance mechanisms, including depth consistency and multi-faceted normal regularization. Extensive evaluations on standard benchmarks demonstrate that our approach achieves state-of-the-art results in novel view synthesis and produces reconstructions with superior geometric accuracy. The results validate that leveraging semantic priors as a geometric regularizer leads to more coherent and complete 3D models from limited input views. Our code is avaliable at \url{https://github.com/MediaX-SJTU/AlignGS}.
\end{abstract}


\section{Introduction}
The advent of radiance fields~\cite{nerf} has revolutionized photorealistic rendering, enabling high quality novel view synthesis for both static~\cite{mipnerf360,3dgs,2dgs,pgsr,sugar} and dynamic scenes~\cite{DeformableGS,hpc,4DGC,jointrf,VRVVC,varfvv}. Among these methods, 3D Gaussian Splatting (3DGS)~\cite{3dgs} has become a particularly prominent approach, prized for its high rendering speed and an editable explicit structure. As applications in augmented reality and robotics increasingly demand high-fidelity, semantically-aware models of our surroundings, applying 3DGS to the task of indoor semantic reconstruction has become a promising and critical area of research.

However, this endeavor faces a dual challenge rooted in the nature of indoor environments. First, due to complex layouts and frequent occlusions, the set of useful photographic views is often sparse. Standard 3DGS pipelines, which rely on classical Structure-from-Motion (SfM)~\cite{colmap}, struggle under these conditions, resulting in flawed reconstructions with distorted and incomplete geometry. Second, to enable semantic understanding, each Gaussian primitive must be endowed with a semantic feature. Existing methods~\cite{GaussianEditor,fearure3dgs,GaussianGrouping,segmentanygaussian} such as Feature 3DGS~\cite{fearure3dgs} and SAGA~\cite{segmentanygaussian} have achieved promising results; however, they implicitly assume the availability of a well-reconstructed geometric foundation onto which features can be projected. This presents a fundamental dilemma: coherent semantic understanding cannot emerge from a flawed or incomplete geometric base.

This paper introduces AlignGS, a novel framework that posits semantic understanding should not be a passive byproduct, but an active, guiding force that regularizes the geometry, especially in the under-constrained sparse-view setting. Our framework realizes this vision by establishing a synergistic, end-to-end joint optimization framework between scene geometry and semantic understanding. Our approach begins with a robust, SfM-free initialization and leverages priors from pre-trained foundation models to actively constrain the geometric optimization through a set of novel guidance mechanisms, namely depth and multi-faceted normal consistency. Extensive evaluations on standard benchmarks, including ScanNet~\cite{scannet} and the NRGBD dataset~\cite{nrgbd}, demonstrate that our approach achieves state-of-the-art results in novel view synthesis and produces reconstructions with superior geometric accuracy. The main contributions of this paper can be summarized as follows:

\begin{itemize}
\item An end-to-end sparse-view indoor reconstruction framework that leverages 2D semantic and geometric priors to directly regularize 3D Gaussian geometry.
\item Semantic-guided constraints—depth consistency and boundary-aware normal consistency—yield more accurate and coherent surfaces from few views.
\item A robust SfM-free initialization via a feed-forward transformer followed by joint optimization, forming a hybrid pipeline with high-fidelity refinement.
\end{itemize}

\section{Method: Semantic-Guided Gaussian Splatting}
\subsection{Overview}
\begin{figure}[htb]
    \centering
    \includegraphics[width=0.95\linewidth]{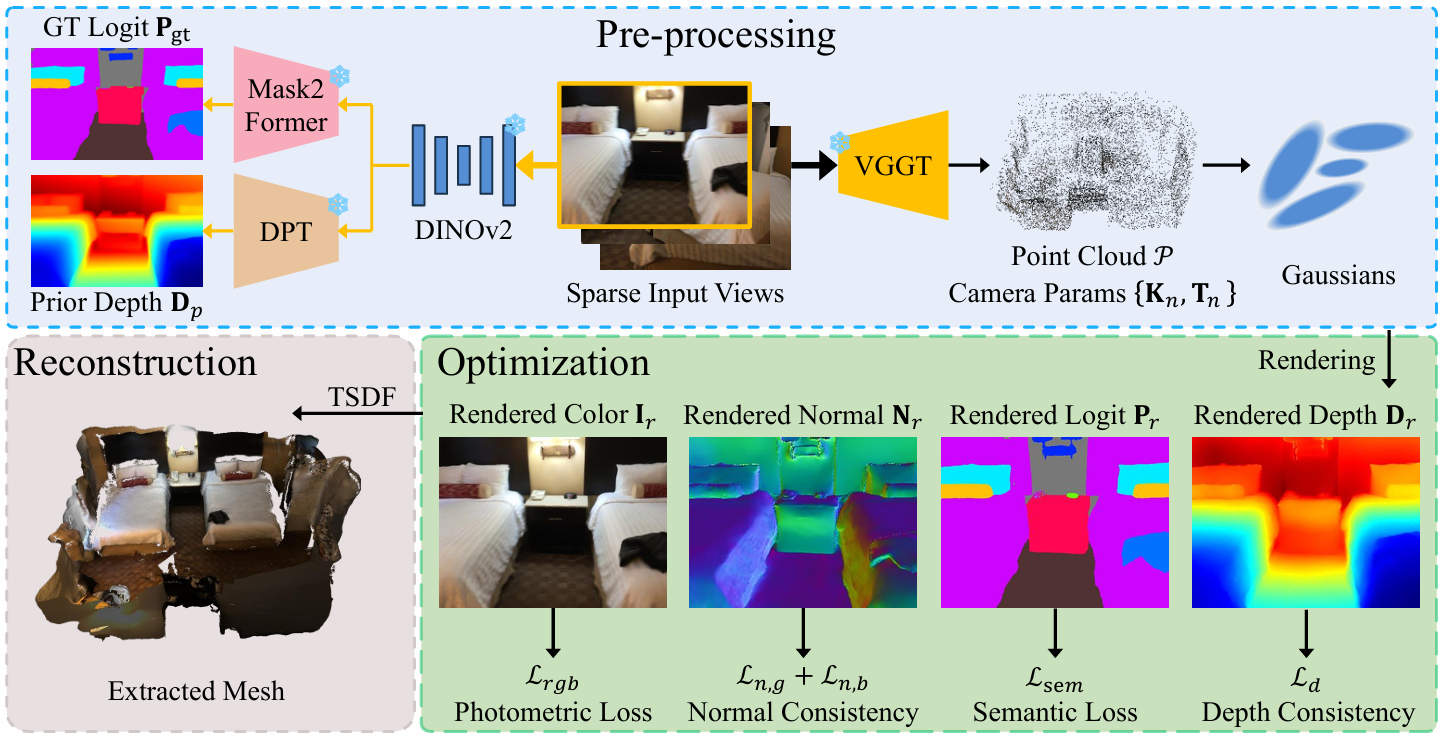}
    \caption{Overview of the AlignGS pipeline, starting with initialization and the subsequent geometry and semantics joint optimization.}
    \label{fig:overview}
\end{figure}
The pipeline of our method is illustrated in Fig.~\ref{fig:overview}. Begin with a sparse set of input images, which are fed into a pre-trained Visual Geometry Grounded Transformer (VGGT) model~\cite{vggt}. This feed-forward network rapidly generates a high-quality initial 3D point cloud, $\mathcal{P}$, and accurate camera poses $\{\mathbf{K}_n, \mathbf{T}_n\}$ without relying on fragile SfM pipelines~\cite{colmap}. Subsequently, we initialize a 3D Gaussian model from this point cloud. Each Gaussian primitive $G_i$ is defined by a set of optimizable parameters $\Theta_i = \{\boldsymbol{\mu}_i, \mathbf{v}_i, \mathbf{q}_i, \alpha_i, \mathbf{c}_i, \mathbf{s}_i\}$, representing its position, scaling vector, rotation quaternion, opacity, color, and semantic vector, respectively. Notably, we augment the standard attributes with a learnable 150-dimensional semantic vector $\mathbf{s}_i \in \mathbb{R}^{150}$. The final stage is a unified optimization loop. Using a differentiable Gaussian renderer, we jointly minimize a set of losses: a photometric reconstruction loss, a semantic loss term, and a semantic-guided geometric constraint loss. This framework enables the end-to-end joint optimization of all geometric and semantic attributes of the Gaussian primitives, ensuring a synergistic refinement of both the scene's geometric structure and its semantic understanding.

\subsection{Lifting 2D Semantics via End-to-End Distillation} \label{sec:semantics-distillation}
To endow our 3D representation with a nuanced understanding of the scene, we leverage a pre-trained 2D teacher model, DINOv2~\cite{dinov2} with a Mask2Former head~\cite{maskformer}, to generate a pseudo-ground-truth semantic logit map $\mathbf{P}_{\text{gt}} \in \mathbb{R}^{H \times W \times C}$ for each input image, where $C=150$. Correspondingly, each gaussian's learnable semantic feature vector $\mathbf{s}_i \in \mathbb{R}^{C}$ can serve as a "student" learning from the teacher's supervisory signal. We distill this knowledge through an end-to-end, dual-supervision strategy. First, for each pixel $u$, we render a semantic logit vector $\mathbf{P}_{r}(u)$ via alpha-blending:
\begin{equation}
    \mathbf{P}_{r}(u) = \sum_{i \in N} \mathbf{s}_i \hat{\alpha_i} \prod_{j=1}^{i-1}(1- \hat{\alpha_j})
    \label{eq:alpha_blend_semantic}
\end{equation}
where $\hat{\alpha_i}$ is the opacity of the projected 2D gaussian. This rendered logit vector is then supervised by the teacher's output $\mathbf{P}_{\text{gt}}(u)$. To this end, we convert both logits to probability distributions, $\hat{\mathbf{P}}_r(u)$ and $\hat{\mathbf{P}}_{\text{gt}}(u)$, using the Softmax function $\sigma(\cdot)$. Our distillation loss, $\mathcal{L}_{\text{sem}}$, then combines two complementary components. 

\noindent \textbf{Soft Distillation} preserves the teacher's nuanced class distribution by minimizing the Kullback-Leibler (KL) divergence:
\begin{equation}
    \mathcal{L}_{\text{soft}} = \frac{1}{|\mathcal{I}|} \sum_{u \in \mathcal{I}} D_{\text{KL}}(\hat{\mathbf{P}}_{\text{gt}}(u) \,||\, \hat{\mathbf{P}}_r(u))
\end{equation}
\textbf{Hard Distillation}, in contrast, enforces alignment with the teacher's most confident prediction. Let $y_{\text{gt}}(u) = \operatorname*{arg\,max}_c (\mathbf{P}_{\text{gt}}(u))_c$ be the hard label from the teacher; the loss is the negative log-probability of this label:
\begin{equation}
    \mathcal{L}_{\text{hard}} = -\frac{1}{|\mathcal{I}|} \sum_{u \in \mathcal{I}} \log\big((\hat{\mathbf{P}}_r(u))_{y_{\text{gt}}(u)}\big)
\end{equation}
The final semantic loss is a weighted combination of these terms, $\mathcal{L}_{\text{sem}} = \lambda_{\text{soft}} \mathcal{L}_{\text{soft}} + \lambda_{\text{hard}} \mathcal{L}_{\text{hard}}$, effectively transferring both ambiguous and deterministic semantic knowledge to the 3D scene representation.

\subsection{Semantic-Guided Geometric Optimization}

This section details our primary contribution: a set of mechanisms that use learned semantic and geometric priors to actively guide and regularize the geometric optimization. This creates a feedback loop where high-level understanding helps refine the 3D shape.

\subsubsection{Depth Consistency Loss}
We regularize the geometry by enforcing consistency between the rendered depth and priors from DINOv2 with a DPT head~\cite{dpt}. However, these monocular depth estimates often suffer from inaccuracies at semantic boundaries, typically caused by occlusions or object discontinuities. To mitigate the impact of these unreliable estimates, we compute our depth loss exclusively on reliable regions. Let $\mathbf{D}_{r}$ and $\mathbf{D}_{p}$ be the rendered and prior depth maps, respectively. We identify pixels corresponding to semantic edges and group them into a mask set $\mathcal{M} \subset \mathcal{I}$, where $\mathcal{I}$ is the set of all pixel coordinates. To handle the inherent scale and shift ambiguity of monocular depth, our loss $\mathcal{L}_{d}$ is formulated using the Pearson correlation coefficient, $\rho$:
\begin{align}
\mathcal{L}_{d} = 1 - \rho(\mathbf{D}_{r}, \mathbf{D}_{p})
\end{align}
where the coefficient $\rho$ is computed exclusively over the reliable (unmasked) pixel set $\mathcal{I} \setminus \mathcal{M}$. This formulation is inherently invariant to scale and shift, and by explicitly excluding pixels in $\mathcal{M}$, we significantly improve the robustness of our geometric supervision.

\subsubsection{Multi-faceted Normal Consistency}
We introduce two normal consistency losses to enforce both geometric self-consistency and semantic coherence. First is \textbf{Geometric Normal Consistency},  we enforce consistency between two distinct sources of surface normals: the map $\mathbf{N}_{r}$, rendered directly from the oriented Gaussian primitives, and the map $\mathbf{N}_{d}$, computed from the spatial gradient of the rendered depth map $\mathbf{D}_{r}$. By minimizing the cosine distance between these two normal fields, we encourage the explicit orientation of the Gaussians to align with the implicit surface they form:
\begin{align}
\mathcal{L}_{n,g} = \frac{1}{|\mathcal{I}|} \sum_{u \in \mathcal{I}} (1 - \mathbf{N}_{r}(u) \cdot \mathbf{N}_{d}(u))    
\end{align}
\noindent \textbf{Semantic Boundary Normal Regularization:} Based on the assumption that geometric surfaces should be discontinuous at the boundaries between different semantic objects, we introduce a loss to promote sharp features. We first identify the set of adjacent pixel pairs that cross a semantic boundary: $\mathcal{B} = \{(u, v) | u \in \mathcal{N}(v), \mathbf{L}_{r}(u) \neq \mathbf{L}_{r}(v)\}$, where $\mathbf{L}_{r}$ is the semantic label map computed from $\mathbf{P}_{\text{gt}}$ using the argmax fuction and $\mathcal{N}(v)$ denotes the neighbors of pixel $v$. 
For each pair in $\mathcal{B}$, we formulate a loss that penalizes similarity and encourages their normals to be divergent:
\begin{align}
\mathcal{L}_{n,b} = \frac{1}{|\mathcal{B}|} \sum_{(u,v)\in \mathcal{B}} \left( 1 - \sigma\big(\alpha \cdot (1 - \mathbf{N}_r(u) \cdot \mathbf{N}_r(v))\big) \right)    
\end{align}
where the term $(1 - \mathbf{N}_r(u) \cdot \mathbf{N}_r(v))$ measures the dissimilarity of the normals. This value is then scaled by a hyperparameter $\alpha$ and passed through a sigmoid function $\sigma(\cdot)$ to produce a bounded loss with stable gradients, ensuring robust optimization.

\subsection{Joint Optimization}
The final stage of our method is the joint optimization of all learnable parameters, driven by a comprehensive loss function $\mathcal{L} = \mathcal{L}_{\text{rgb}} + \lambda_{\text{sem}} \mathcal{L}_{\text{sem}} + \lambda_{\text{guide}} \mathcal{L}_{\text{guide}}$,
where $\lambda_{\text{sem}}$ and $\lambda_{\text{guide}}$ are hyperparameters balancing the main components. The photometric loss $\mathcal{L}_{\text{rgb}}$ is a standard combination of an L1 loss and a structural similarity (D-SSIM) term:
    \begin{equation}
    \mathcal{L}_{\text{rgb}} = (1-\lambda_{\text{ssim}})\|\mathbf{I}_{\text{gt}} - \mathbf{I}_{r}\|_1 + \lambda_{\text{ssim}}(1 - \text{SSIM}(\mathbf{I}_{\text{gt}}, \mathbf{I}_{r}))
    \end{equation}
    where $\mathbf{I}_{\text{gt}}$ and $\mathbf{I}_{r}$ are the ground-truth and rendered images.
The semantic loss $\mathcal{L}_{\text{sem}}$ is our dual-supervision distillation term  detailed in \ref{sec:semantics-distillation}.
The geometric guidance loss $\mathcal{L}_{\text{guide}}$ aggregates our regularizers, weighted by their respective hyperparameters $\omega$: $\mathcal{L}_{\text{guide}} = \omega_{d} \mathcal{L}_d + \omega_{ng} \mathcal{L}_{n,g} + \omega_{nb} \mathcal{L}_{n,b}$. 
By minimizing the total loss $\mathcal{L}$, our framework ensures that geometric refinement and semantic understanding proceed in synergy.

\section{Experiments}
\subsection{Implementation Details}
We build on PGSR~\cite{pgsr} and train for 7k iterations on a single RTX 4090. Unless stated, all hyperparameters are fixed across experiments: $\lambda_{\text{sem}}{=}1.0$, $\lambda_{\text{guide}}{=}1.0$, $\lambda_{\text{soft}}{=}1.0$, $\lambda_{\text{hard}}{=}0.1$, $\lambda_{\text{ssim}}{=}0.2$, $\omega_{d}{=}0.5$, $\omega_{ng}{=}0.05$, $\omega_{nb}{=}0.01$, $\alpha{=}100$. Semantic-guided geometry is enabled after 1.5k iters. Meshes are extracted via TSDF fusion~\cite{neus2}. We evaluate on 6 ScanNet~\cite{scannet} and 4 NRGBD~\cite{nrgbd} scenes at $640{\times}480$ with sparse training views (ScanNet: 36/1383; NRGBD: 20/1185).

\subsection{Comparison}
\subsubsection{Quantitative Comparisons}
To evaluate the performance of our proposed method, we conduct comprehensive comparisons against a range of state-of-the-art methods including Neuralangelo~\cite{neuralangelo}, 3DGS~\cite{3dgs}, 2DGS~\cite{2dgs}, PGSR~\cite{pgsr}, FSGS~\cite{FSGS} and SparseGS~\cite{sparsegs}. For these baselines, we adopt their default training strategey and hyperparameters and initialize with COLMAP point cloud. As shown in Tab.~\ref{tab:all_nvs_compare}, our method, AlignGS, demonstrates state-of-the-art performance in novel view synthesis across both the real-world ScanNet and synthetic NRGBD datasets. On ScanNet, AlignGS achieves a PSNR of 25.93 on \texttt{scene0085\_00}, significantly outperforming all baselines. Such strong performance also extends to the synthetic domain.
This trend of superior performance, particularly in the LPIPS metric which indicates higher perceptual quality, is consistent across the vast majority of tested scenes. In terms of geometric fidelity (Tab.~\ref{tab:all_mesh_compare}), AlignGS produces substantially more accurate and complete meshes. Our method consistently achieves the highest F-score across both datasets, indicating a superior balance of precision and recall. On the ScanNet scene \texttt{scene0625\_00}, our method achieves an F-score of 0.601, nearly doubling the next best result from 2DGS (0.304). Similarly on the NRGBD 
, AlignGS 
again obtain a significant improvement over all other methods. This robust geometric improvement highlights the effectiveness of our semantic-guided constraints in producing accurate and coherent surfaces from sparse views.

\begin{table*}[htb]
\setlength{\tabcolsep}{2.5pt} 
\renewcommand{\arraystretch}{1.05}
\centering
\caption{Quantitative novel view synthesis comparison on ScanNet and NRGBD datasets.}
\vspace{-2mm}
\label{tab:all_nvs_compare}
\resizebox{\textwidth}{!}{%
\begin{tabular}{@{}l|ccc|ccc|ccc|ccc|ccc@{}}
\hline
\multirow{2}{*}{Method} & \multicolumn{3}{c|}{\texttt{scene0085\_00}} & \multicolumn{3}{c|}{\texttt{scene0009\_01}} & \multicolumn{3}{c|}{\texttt{scene0114\_02}} & \multicolumn{3}{c|}{\texttt{scene0617\_00}} & \multicolumn{3}{c}{\texttt{scene0625\_00}} \\
& PSNR$\uparrow$ & SSIM$\uparrow$ & LPIPS$\downarrow$ & PSNR$\uparrow$ & SSIM$\uparrow$ & LPIPS$\downarrow$ & PSNR$\uparrow$ & SSIM$\uparrow$ & LPIPS$\downarrow$ & PSNR$\uparrow$ & SSIM$\uparrow$ & LPIPS$\downarrow$ & PSNR$\uparrow$ & SSIM$\uparrow$ & LPIPS$\downarrow$ \\
\hline
Neuralangelo &15.02&0.610&0.446 & 16.16 & 0.676& 0.544 & 12.09 & 0.587 & 0.653 & 11.07 & 0.460 & 0.720 & 21.59 & \cellcolor{lightorange}0.803 & \cellcolor{lightorange}0.357 \\
3DGS     & 16.94 & 0.617 & 0.455 & 19.88 & 0.615 & 0.413 & 15.54 & 0.597 & 0.506 & 15.66 & 0.494 & 0.528 & 20.55 & 0.714 & 0.447 \\
2DGS     & 17.48 & 0.664 & 0.414 & 16.77 & 0.616 & 0.441 & 14.98 & 0.614 & 0.475 & 15.25 & 0.517 & 0.513 & 23.11 & 0.763 & 0.390 \\
PGSR     & 18.30 & 0.664 & 0.412 & 16.96 & 0.552 & 0.456 & 11.67 & 0.446 & 0.581 & 15.64 & 0.485 & 0.521 & 16.52 & 0.654 & 0.495 \\
FSGS  & 20.25 & \cellcolor{lightorange}0.731 & 0.395 & \cellcolor{lightorange}20.83 & \cellcolor{lightorange}0.684 & \cellcolor{lightorange}0.394 & \cellcolor{lightorange}19.02 & \cellcolor{lightorange}0.711 & 0.436 & 16.56 & \cellcolor{lightorange}0.575 & 0.489 & \cellcolor{lightorange}24.74 & \cellcolor{lightred}0.807 & 0.377 \\
SparseGS  & \cellcolor{lightorange}20.29 & 0.693 & \cellcolor{lightorange}0.381 & 17.61 & 0.625 & 0.404 & 18.03 & 0.670 & \cellcolor{lightorange}0.434 & \cellcolor{lightorange}16.75 & 0.548 & \cellcolor{lightorange}0.483 & 23.14 & 0.750 &0.377 \\
Ours         & \cellcolor{lightred}25.93 & \cellcolor{lightred}0.833 & \cellcolor{lightred}0.283 & \cellcolor{lightred}22.41 & \cellcolor{lightred}0.691 & \cellcolor{lightred}0.335 & \cellcolor{lightred}21.53 & \cellcolor{lightred}0.780 & \cellcolor{lightred}0.371 & \cellcolor{lightred}22.02 & \cellcolor{lightred}0.716 & \cellcolor{lightred}0.357 & \cellcolor{lightred}25.31 & \cellcolor{lightorange}0.803 & \cellcolor{lightred}0.334 \\
\hline
& \multicolumn{3}{c|}{\texttt{scene0771\_00}} & \multicolumn{3}{c|}{\texttt{complete\_kitchen}} & \multicolumn{3}{c|}{\texttt{breakfast\_room}} & \multicolumn{3}{c|}{\texttt{green\_room}} & \multicolumn{3}{c}{\texttt{morning\_apartment}} \\
\hline
Neuralangelo     & 16.99 & \cellcolor{lightorange}0.695 & 0.523 & 17.22 & 0.580 & 0.660 & 22.10 & 0.821 & 0.272 & 14.93 & 0.681 & 0.651 & 14.45 & 0.662 & 0.514 \\
3DGS     & 16.27 & 0.572 & 0.519 & 23.23 & 0.778 & 0.258 & 26.70 & 0.936 & 0.103 & 25.06 & 0.902 & 0.139 & 27.51 & 0.896 & 0.140 \\
2DGS     & 15.94 & 0.626 & 0.502 & 25.02 & 0.861 & 0.215 & 21.61 & 0.907 & 0.175 & 26.96 & 0.891 & 0.150 & 28.38 & 0.911 & 0.123 \\
PGSR     & 19.10 & 0.634 & 0.469 & 23.46 & 0.809 & 0.247 & 26.88 & 0.912 & 0.175 & 23.56 & 0.868 & 0.164 & \cellcolor{lightred}28.81 & 0.912 & \cellcolor{lightorange}0.121 \\
FSGS  & 19.12 & \cellcolor{lightorange}0.695 & 0.491 & 25.29 & 0.854 & 0.294 & 28.74 & 0.945 & 0.106 & \cellcolor{lightorange}29.32 & \cellcolor{lightred}0.928 & \cellcolor{lightorange}0.106 & 28.31 & \cellcolor{lightorange}0.915 & 0.123 \\
SparseGS  & \cellcolor{lightorange}19.30 & 0.680 & \cellcolor{lightorange}0.447 & \cellcolor{lightorange}26.40 & \cellcolor{lightorange}0.867 & \cellcolor{lightorange}0.200 & \cellcolor{lightorange}31.22 & \cellcolor{lightorange}0.951 & \cellcolor{lightorange}0.083 & 29.10 & 0.924 & 0.114 & 27.89 & 0.905 & 0.137 \\
Ours         & \cellcolor{lightred}24.25 & \cellcolor{lightred}0.764 & \cellcolor{lightred}0.372 & \cellcolor{lightred}27.18 & \cellcolor{lightred}0.876 & \cellcolor{lightred}0.193 & \cellcolor{lightred}31.34 & \cellcolor{lightred}0.953 & \cellcolor{lightred}0.074 & \cellcolor{lightred}29.91 & \cellcolor{lightorange}0.927 & \cellcolor{lightred}0.088 & \cellcolor{lightorange}28.74 & \cellcolor{lightred}0.916 & \cellcolor{lightred}0.109 \\
\hline
\end{tabular}%
}
\end{table*}

\begin{table*}[htb]
\vspace{-3mm}
\setlength{\tabcolsep}{2pt}
\renewcommand{\arraystretch}{1.05}
\centering
\caption{Geometry quality comparison on ScanNet and NRGBD datasets.}
\label{tab:all_mesh_compare}
\vspace{-2mm}
\resizebox{\textwidth}{!}{%
\begin{tabular}{@{}l|ccccc|ccccc|ccccc|ccccc|ccccc@{}}
\hline
\multirow{2}{*}{Method} & \multicolumn{5}{c|}{\texttt{scene0085\_00}} & \multicolumn{5}{c|}{\texttt{scene0009\_01}} & \multicolumn{5}{c|}{\texttt{scene0114\_02}} & \multicolumn{5}{c|}{\texttt{scene0617\_00}} & \multicolumn{5}{c}{\texttt{scene0625\_00}} \\
& Acc$\downarrow$ & Comp$\downarrow$ & Prec$\uparrow$ & Recall$\uparrow$ & F-score$\uparrow$ & Acc$\downarrow$ & Comp$\downarrow$ & Prec$\uparrow$ & Recall$\uparrow$ & F-score$\uparrow$ & Acc$\downarrow$ & Comp$\downarrow$ & Prec$\uparrow$ & Recall$\uparrow$ & F-score$\uparrow$ & Acc$\downarrow$ & Comp$\downarrow$ & Prec$\uparrow$ & Recall$\uparrow$ & F-score$\uparrow$ & Acc$\downarrow$ & Comp$\downarrow$ & Prec$\uparrow$ & Recall$\uparrow$ & F-score$\uparrow$ \\
\hline
3DGS    & 0.383 & 0.872 & 0.082 & 0.006 & 0.012 & 0.412 & 0.775 & 0.027 & 0.003 & 0.005 & 0.419 & 0.543 & 0.096 & 0.014 & 0.024 & 0.300 & 0.687 & 0.094 & 0.008 & 0.015 & 0.132 & 0.776 & 0.155 & 0.014 & 0.026 \\
2DGS  & 0.153 & \cellcolor{lightorange}0.140 & 0.292 & \cellcolor{lightorange}0.292 & 0.292 & \cellcolor{lightorange}0.185 & \cellcolor{lightorange}0.149 & 0.156 & \cellcolor{lightorange}0.175 & \cellcolor{lightorange}0.165 & \cellcolor{lightorange}0.198 & \cellcolor{lightorange}0.233 & \cellcolor{lightorange}0.251 & \cellcolor{lightorange}0.156 & \cellcolor{lightorange}0.192 & 0.135 & \cellcolor{lightorange}0.108 & \cellcolor{lightorange}0.356 & \cellcolor{lightorange}0.347 & \cellcolor{lightorange}0.351 & \cellcolor{lightorange}0.124 & \cellcolor{lightorange}0.142 & \cellcolor{lightorange}0.318 & \cellcolor{lightorange}0.290 & \cellcolor{lightorange}0.304 \\
PGSR     & \cellcolor{lightorange}0.139 & 0.148 & \cellcolor{lightorange}0.310 & 0.290 & \cellcolor{lightorange}0.300 & 0.230 & 0.189 & \cellcolor{lightorange}0.158 & 0.171 & 0.164 & 0.237 & 0.333 & 0.198 & 0.112 & 0.143 & \cellcolor{lightorange}0.133 & 0.118 & 0.309 & 0.302 & 0.305 & 0.155 & 0.182 & 0.236 & 0.190 & 0.210 \\
Ours         & \cellcolor{lightred}0.055 & \cellcolor{lightred}0.065 & \cellcolor{lightred}0.562 & \cellcolor{lightred}0.505 & \cellcolor{lightred}0.532 & \cellcolor{lightred}0.125 & \cellcolor{lightred}0.148 & \cellcolor{lightred}0.324 & \cellcolor{lightred}0.273 & \cellcolor{lightred}0.297 & \cellcolor{lightred}0.123 & \cellcolor{lightred}0.109 & \cellcolor{lightred}0.310 & \cellcolor{lightred}0.292 & \cellcolor{lightred}0.301 & \cellcolor{lightred}0.079 & \cellcolor{lightred}0.060 & \cellcolor{lightred}0.541 & \cellcolor{lightred}0.575 & \cellcolor{lightred}0.557 & \cellcolor{lightred}0.050 & \cellcolor{lightred}0.049 & \cellcolor{lightred}0.594 & \cellcolor{lightred}0.607 & \cellcolor{lightred}0.601 \\
\hline
& \multicolumn{5}{c|}{\texttt{scene0771\_00}} & \multicolumn{5}{c|}{\texttt{complete\_kitchen}} & \multicolumn{5}{c|}{\texttt{breakfast\_room}} & \multicolumn{5}{c|}{\texttt{green\_room}} & \multicolumn{5}{c}{\texttt{morning\_apartment}} \\
\hline
3DGS      & 0.291 & 0.770 & 0.093 & 0.010 & 0.018 & 0.678 & 1.587 & 0.027 & 0.002 & 0.003 & 0.616 & 0.650 & 0.040 & 0.011 & 0.017 & 0.519 & 0.860 & 0.058 & 0.015 & 0.024 & 0.463 & 0.629 & 0.082 & 0.033 & 0.047 \\
2DGS    & \cellcolor{lightorange}0.227 & \cellcolor{lightorange}0.266 & 0.178 & \cellcolor{lightorange}0.136 & \cellcolor{lightorange}0.154 & 0.182 & 0.610 & 0.411 & 0.295 & 0.344 & 0.236 & 0.355 & \cellcolor{lightorange}0.322 & 0.221 & \cellcolor{lightorange}0.262 & \cellcolor{lightorange}0.142 & \cellcolor{lightorange}0.198 & 0.411 & 0.305 & 0.350 & 0.101 & 0.150 & 0.517 & 0.384 & 0.440 \\
PGSR     & 0.248 & 0.352 & \cellcolor{lightorange}0.199 & 0.117 & 0.148 & \cellcolor{lightorange}0.171 & \cellcolor{lightorange}0.452 & \cellcolor{lightred}0.523 & \cellcolor{lightorange}0.404 & \cellcolor{lightorange}0.456 & \cellcolor{lightorange}0.194 & \cellcolor{lightorange}0.309 & 0.307 & \cellcolor{lightorange}0.223 & 0.258 & 0.172 & 0.211 & \cellcolor{lightorange}0.414 & \cellcolor{lightorange}0.334 & \cellcolor{lightorange}0.370 & \cellcolor{lightorange}0.093 & \cellcolor{lightorange}0.124 & \cellcolor{lightorange}0.593 & \cellcolor{lightorange}0.477 & \cellcolor{lightorange}0.529 \\
Ours         & \cellcolor{lightred}0.111 & \cellcolor{lightred}0.124 & \cellcolor{lightred}0.377 & \cellcolor{lightred}0.313 & \cellcolor{lightred}0.342 & \cellcolor{lightred}0.121 & \cellcolor{lightred}0.152 & \cellcolor{lightorange}0.495 & \cellcolor{lightred}0.466 & \cellcolor{lightred}0.480 & \cellcolor{lightred}0.086 & \cellcolor{lightred}0.146 & \cellcolor{lightred}0.397 & \cellcolor{lightred}0.380 & \cellcolor{lightred}0.388 & \cellcolor{lightred}0.075 & \cellcolor{lightred}0.088 & \cellcolor{lightred}0.421 & \cellcolor{lightred}0.378 & \cellcolor{lightred}0.398 & \cellcolor{lightred}0.039 & \cellcolor{lightred}0.063 & \cellcolor{lightred}0.737 & \cellcolor{lightred}0.616 & \cellcolor{lightred}0.671 \\
\hline
\end{tabular}%
}
\vspace{-5mm}
\end{table*}

\subsubsection{Qualitative Comparisons}

\begin{figure}[htbp]
  \centering
  \includegraphics[width=0.95\linewidth]{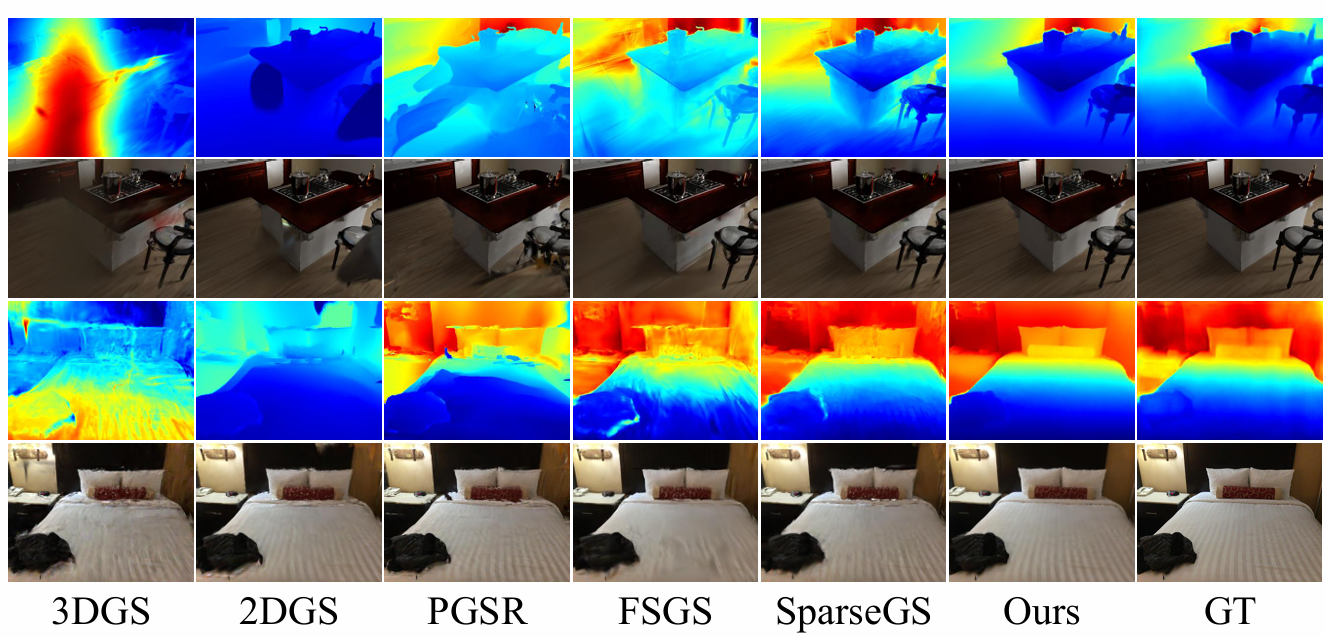} 
  \vspace{-3mm}
  \caption{Qualitative NVS comparisons across ScanNet and NRGBD scenes.}
  \label{fig:qualitative-nvs-compare}
\end{figure}

\begin{figure}[htbp]
  \centering
  \includegraphics[width=0.95\linewidth]{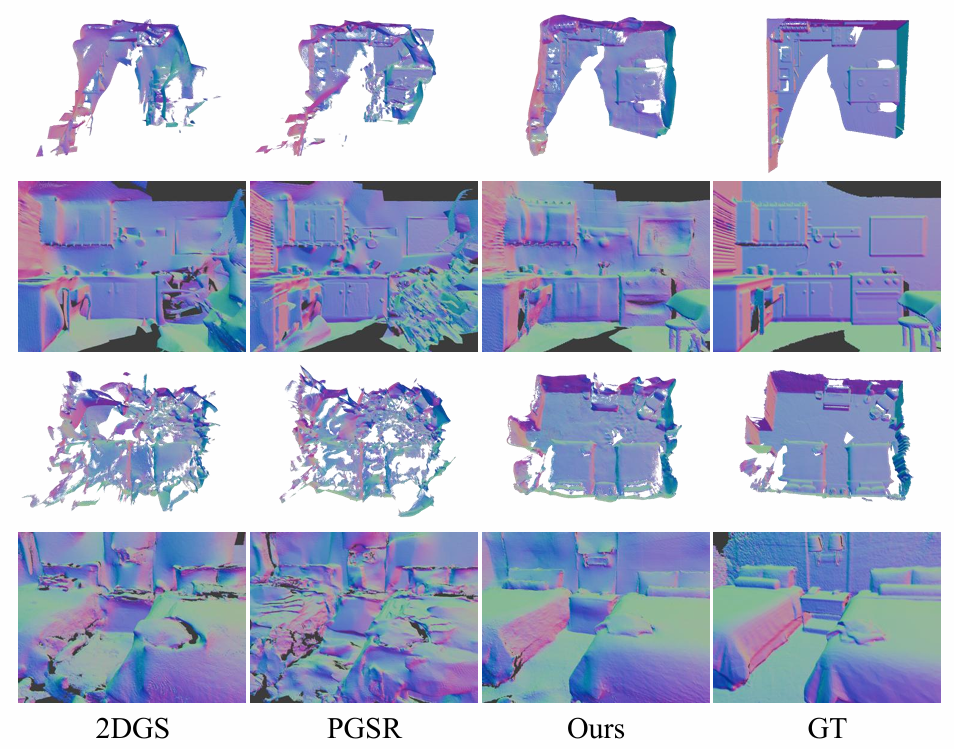} 
  \vspace{-3mm}
  \caption{Qualitative reconstruction comparisons across ScanNet and NRGBD scenes.}
  \label{fig:mesh-compare}
\end{figure}

\begin{figure}[htb]
    \centering
    \includegraphics[width=0.95\linewidth]{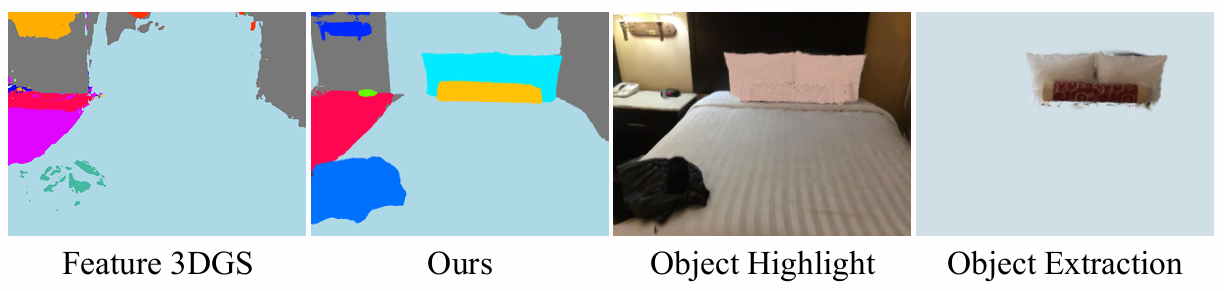}
    \vspace{-2mm}
    \caption{Semantic Applications}
    \label{fig:semantic-edit}
\end{figure}

We present a qualitative comparison in Fig.~\ref{fig:qualitative-nvs-compare} and Fig.~\ref{fig:mesh-compare}, showing our rendered RGB and depth from novel viewpoints, as well as mesh normal alongside key baselines and ground truth. Our method produces significantly fewer artifacts and exhibits more coherent geometric structures compared to others. Furthermore, our method recovers structurally more accurate and higher-fidelity surfaces with improved smoothness on objects and sharper distinction at semantic boundaries, thanks to the SfM-free initialization followed by the semantic-guided geometric regularization.  And we present our downstream editing performance in Fig.~\ref{fig:semantic-edit}. The left two columns compare our segmentation results from a novel viewpoint with Feature 3DGS. Leveraging robust semantic masks from DINOv2 and dual semantic supervision, our method achieves more accurate and complete semantic boundaries. The right two columns demonstrate language-guided editing, including object extraction, deletion, and color highlighting (e.g., for pillows and cushions).

\subsection{Ablation Study}
We conduct a progressive ablation study on the \texttt{morning\_apartment} scene, start with a base model built upon the PGSR framework. As shown in Tab.~\ref{tab:ablation_morning_apartment}, Starting from a PGSR base, replacing SfM with VGGT notably strengthens geometry (F-score $\uparrow$ 0.464$\to$0.565). Building on this, adding $+ \mathcal{L}_{sem}$ further enhances geometric completeness (F-score to 0.570) while having a minor positive impact on rendering quality. The subsequent addition of our depth consistency loss ($+ \mathcal{L}_{d}$) yields a substantial leap in geometric accuracy (F-score to 0.618) and steadily improves the PSNR to 28.05. Finally, incorporating the multi-faceted normal consistency losses ($+ \mathcal{L}_{n,g}+\mathcal{L}_{n,b}$) provides the last and most significant boost. This component drastically improves rendering quality, with PSNR jumping to 28.74, while also pushing geometric fidelity to its best value (F-score of 0.671). This analysis confirms that each module in our AlignGS framework provides a distinct and cumulative benefit to both geometry and appearance.

\vspace{-4mm}
\begin{table}[htb]
\setlength{\tabcolsep}{2pt} 
\renewcommand{\arraystretch}{1.05}
\centering
\caption{Ablation study conducted on the morning apartment scene from the NRGBD dataset, each module is cumulatively added.}
\label{tab:ablation_morning_apartment}
\vspace{-2mm}
\resizebox{0.98\linewidth}{!}{%
\begin{tabular}{l|ccc|ccccc}
\hline
Method & PSNR$\uparrow$ & SSIM$\uparrow$ & LPIPS$\downarrow$ & Acc$\downarrow$ & Comp$\downarrow$ & Prec$\uparrow$ & Recall$\uparrow$ & F-score$\uparrow$ \\
\hline
base & 27.38 & 0.903 & 0.127 & 0.101 & 0.146 & 0.531 & 0.411 & 0.464 \\
$+$ VGGT & 27.85 & 0.908 & 0.123 & 0.074 & 0.103 & 0.609 & 0.527 & 0.565 \\
$+ \mathcal{L}_{sem}$ & 27.91 & 0.908 & 0.122 & 0.067 & 0.084 & 0.603 & 0.541 & 0.570 \\
$+ \mathcal{L}_{d}$ & 28.05 & 0.905 & 0.129 & 0.046 & 0.069 & 0.665 & 0.577 & 0.618 \\
$+ \mathcal{L}_{n,g}+\mathcal{L}_{n,b} $  & 28.74 & 0.916 & 0.109 & 0.039 & 0.063 & 0.737 & 0.616 & 0.671 \\
\hline
\end{tabular}
}
\end{table}

\vspace{-5mm}
\section{Conclusion and Discussion}
In this paper, we present AlignGS, a novel framework for semantic 3D indoor reconstruction from sparse views. Unlike prior methods that project semantics onto fixed geometry, AlignGS integrates semantic priors from 2D foundation models to directly guide geometric optimization in an end-to-end manner. This synergy resolves geometric ambiguities in ill-posed settings and achieves state-of-the-art performance in both novel view synthesis and geometric fidelity. Moreover, the explicit per-primitive semantic features learned by our model enable downstream tasks such as object editing and asset replacement, advancing the creation of high-fidelity, semantically-aware digital twins.

\clearpage
\bibliographystyle{IEEEtran}
\bibliography{reference}

\begin{thebibliography}{10}
\providecommand{\url}[1]{#1}
\csname url@samestyle\endcsname
\providecommand{\newblock}{\relax}
\providecommand{\bibinfo}[2]{#2}
\providecommand{\BIBentrySTDinterwordspacing}{\spaceskip=0pt\relax}
\providecommand{\BIBentryALTinterwordstretchfactor}{4}
\providecommand{\BIBentryALTinterwordspacing}{\spaceskip=\fontdimen2\font plus
\BIBentryALTinterwordstretchfactor\fontdimen3\font minus \fontdimen4\font\relax}
\providecommand{\BIBforeignlanguage}[2]{{%
\expandafter\ifx\csname l@#1\endcsname\relax
\typeout{** WARNING: IEEEtran.bst: No hyphenation pattern has been}%
\typeout{** loaded for the language `#1'. Using the pattern for}%
\typeout{** the default language instead.}%
\else
\language=\csname l@#1\endcsname
\fi
#2}}
\providecommand{\BIBdecl}{\relax}
\BIBdecl

\bibitem{nerf}
B.~Mildenhall, P.~P. Srinivasan, M.~Tancik, J.~T. Barron, R.~Ramamoorthi, and R.~Ng, ``Nerf: representing scenes as neural radiance fields for view synthesis,'' \emph{Commun. ACM}, vol.~65, no.~1, p. 99–106, December 2021.

\bibitem{mipnerf360}
J.~T. Barron, B.~Mildenhall, D.~Verbin, P.~P. Srinivasan, and P.~Hedman, ``Mip-nerf 360: Unbounded anti-aliased neural radiance fields,'' in \emph{2022 IEEE/CVF Conference on Computer Vision and Pattern Recognition (CVPR)}, 2022, pp. 5460--5469.

\bibitem{3dgs}
B.~Kerbl, G.~Kopanas, T.~Leimkuehler, and G.~Drettakis, ``3d gaussian splatting for real-time radiance field rendering,'' \emph{ACM Trans. Graph.}, vol.~42, no.~4, July 2023.

\bibitem{2dgs}
B.~Huang, Z.~Yu, A.~Chen, A.~Geiger, and S.~Gao, ``2d gaussian splatting for geometrically accurate radiance fields,'' in \emph{ACM SIGGRAPH 2024 Conference Papers}, ser. SIGGRAPH '24.\hskip 1em plus 0.5em minus 0.4em\relax New York, NY, USA: Association for Computing Machinery, 2024.

\bibitem{pgsr}
D.~Chen, H.~Li, W.~Ye, Y.~Wang, W.~Xie, S.~Zhai, N.~Wang, H.~Liu, H.~Bao, and G.~Zhang, ``Pgsr: Planar-based gaussian splatting for efficient and high-fidelity surface reconstruction,'' \emph{IEEE Transactions on Visualization and Computer Graphics}, pp. 1--12, 2024.

\bibitem{sugar}
A.~Guédon and V.~Lepetit, ``Sugar: Surface-aligned gaussian splatting for efficient 3d mesh reconstruction and high-quality mesh rendering,'' in \emph{2024 IEEE/CVF Conference on Computer Vision and Pattern Recognition (CVPR)}, 2024, pp. 5354--5363.

\bibitem{DeformableGS}
Z.~Yang, X.~Gao, W.~Zhou, S.~Jiao, Y.~Zhang, and X.~Jin, ``Deformable 3d gaussians for high-fidelity monocular dynamic scene reconstruction,'' in \emph{2024 IEEE/CVF Conference on Computer Vision and Pattern Recognition (CVPR)}, 2024, pp. 20\,331--20\,341.

\bibitem{hpc}
Z.~Zheng, H.~Zhong, Q.~Hu, X.~Zhang, L.~Song, Y.~Zhang, and Y.~Wang, ``Hpc: Hierarchical progressive coding framework for volumetric video,'' in \emph{Proceedings of the 32nd ACM International Conference on Multimedia}, 2024, pp. 7937--7946.

\bibitem{4DGC}
Q.~Hu, Z.~Zheng, H.~Zhong, S.~Fu, L.~Song, X.~Zhang, G.~Zhai, and Y.~Wang, ``4dgc: Rate-aware 4d gaussian compression for efficient streamable free-viewpoint video,'' in \emph{Proceedings of the Computer Vision and Pattern Recognition Conference (CVPR)}, June 2025, pp. 875--885.

\bibitem{jointrf}
Z.~Zheng, H.~Zhong, Q.~Hu, X.~Zhang, L.~Song, Y.~Zhang, and Y.~Wang, ``Jointrf: End-to-end joint optimization for dynamic neural radiance field representation and compression,'' in \emph{2024 IEEE International Conference on Image Processing (ICIP)}, 2024, pp. 3292--3298.

\bibitem{VRVVC}
Q.~Hu, H.~Zhong, Z.~Zheng, X.~Zhang, Z.~Cheng, L.~Song, G.~Zhai, and Y.~Wang, ``Vrvvc: Variable-rate nerf-based volumetric video compression,'' in \emph{Proceedings of the AAAI Conference on Artificial Intelligence}, vol.~39, no.~4, 2025, pp. 3563--3571.

\bibitem{varfvv}
Q.~Hu, Q.~He, H.~Zhong, G.~Lu, X.~Zhang, G.~Zhai, and Y.~Wang, ``Varfvv: View-adaptive real-time interactive free-view video streaming with edge computing,'' \emph{IEEE Journal on Selected Areas in Communications}, vol.~43, no.~7, pp. 2620--2634, 2025.

\bibitem{colmap}
J.~L. Schönberger and J.-M. Frahm, ``Structure-from-motion revisited,'' in \emph{2016 IEEE Conference on Computer Vision and Pattern Recognition (CVPR)}, 2016, pp. 4104--4113.

\bibitem{GaussianEditor}
Y.~Chen, Z.~Chen, C.~Zhang, F.~Wang, X.~Yang, Y.~Wang, Z.~Cai, L.~Yang, H.~Liu, and G.~Lin, ``Gaussianeditor: Swift and controllable 3d editing with gaussian splatting,'' in \emph{Proceedings of the IEEE/CVF Conference on Computer Vision and Pattern Recognition (CVPR)}, June 2024, pp. 21\,476--21\,485.

\bibitem{fearure3dgs}
S.~Zhou, H.~Chang, S.~Jiang, Z.~Fan, Z.~Zhu, D.~Xu, P.~Chari, S.~You, Z.~Wang, and A.~Kadambi, ``Feature 3dgs: Supercharging 3d gaussian splatting to enable distilled feature fields,'' in \emph{Proceedings of the IEEE/CVF Conference on Computer Vision and Pattern Recognition (CVPR)}, June 2024, pp. 21\,676--21\,685.

\bibitem{GaussianGrouping}
M.~Ye, M.~Danelljan, F.~Yu, and L.~Ke, ``Gaussian grouping: Segment and edit anything in 3d scenes,'' in \emph{Computer Vision -- ECCV 2024}, A.~Leonardis, E.~Ricci, S.~Roth, O.~Russakovsky, T.~Sattler, and G.~Varol, Eds.\hskip 1em plus 0.5em minus 0.4em\relax Cham: Springer Nature Switzerland, 2025, pp. 162--179.

\bibitem{segmentanygaussian}
J.~Cen, J.~Fang, C.~Yang, L.~Xie, X.~Zhang, W.~Shen, and Q.~Tian, ``Segment any 3d gaussians,'' in \emph{Proceedings of the AAAI Conference on Artificial Intelligence}, vol.~39, no.~2, 2025, pp. 1971--1979.

\bibitem{scannet}
A.~Dai, A.~X. Chang, M.~Savva, M.~Halber, T.~Funkhouser, and M.~Nießner, ``Scannet: Richly-annotated 3d reconstructions of indoor scenes,'' in \emph{2017 IEEE Conference on Computer Vision and Pattern Recognition (CVPR)}, 2017, pp. 2432--2443.

\bibitem{nrgbd}
D.~Azinović, R.~Martin-Brualla, D.~B. Goldman, M.~Nießner, and J.~Thies, ``Neural rgb-d surface reconstruction,'' in \emph{2022 IEEE/CVF Conference on Computer Vision and Pattern Recognition (CVPR)}, 2022, pp. 6280--6291.

\bibitem{vggt}
J.~Wang, M.~Chen, N.~Karaev, A.~Vedaldi, C.~Rupprecht, and D.~Novotny, ``Vggt: Visual geometry grounded transformer,'' in \emph{Proceedings of the Computer Vision and Pattern Recognition Conference (CVPR)}, June 2025, pp. 5294--5306.

\bibitem{dinov2}
M.~Oquab, T.~Darcet, T.~Moutakanni, H.~V. Vo, M.~Szafraniec, V.~Khalidov, P.~Fernandez, D.~HAZIZA, F.~Massa, A.~El-Nouby, M.~Assran, N.~Ballas, W.~Galuba, R.~Howes, P.-Y. Huang, S.-W. Li, I.~Misra, M.~Rabbat, V.~Sharma, G.~Synnaeve, H.~Xu, H.~Jegou, J.~Mairal, P.~Labatut, A.~Joulin, and P.~Bojanowski, ``Dinov2: Learning robust visual features without supervision,'' \emph{Transactions on Machine Learning Research}, 2024.

\bibitem{maskformer}
B.~Cheng, I.~Misra, A.~G. Schwing, A.~Kirillov, and R.~Girdhar, ``Masked-attention mask transformer for universal image segmentation,'' in \emph{Proceedings of the IEEE/CVF Conference on Computer Vision and Pattern Recognition (CVPR)}, June 2022, pp. 1290--1299.

\bibitem{dpt}
R.~Ranftl, A.~Bochkovskiy, and V.~Koltun, ``Vision transformers for dense prediction,'' in \emph{Proceedings of the IEEE/CVF International Conference on Computer Vision (ICCV)}, October 2021, pp. 12\,179--12\,188.

\bibitem{neus2}
Y.~Wang, Q.~Han, M.~Habermann, K.~Daniilidis, C.~Theobalt, and L.~Liu, ``Neus2: Fast learning of neural implicit surfaces for multi-view reconstruction,'' in \emph{2023 IEEE/CVF International Conference on Computer Vision (ICCV)}, 2023, pp. 3272--3283.

\bibitem{neuralangelo}
Z.~Li, T.~M\"uller, A.~Evans, R.~H. Taylor, M.~Unberath, M.-Y. Liu, and C.-H. Lin, ``Neuralangelo: High-fidelity neural surface reconstruction,'' in \emph{Proceedings of the IEEE/CVF Conference on Computer Vision and Pattern Recognition (CVPR)}, June 2023, pp. 8456--8465.

\bibitem{FSGS}
Z.~Zhu, Z.~Fan, Y.~Jiang, and Z.~Wang, ``Fsgs: Real-time few-shot view synthesis using gaussian splatting,'' in \emph{Computer Vision -- ECCV 2024}, A.~Leonardis, E.~Ricci, S.~Roth, O.~Russakovsky, T.~Sattler, and G.~Varol, Eds.\hskip 1em plus 0.5em minus 0.4em\relax Cham: Springer Nature Switzerland, 2025, pp. 145--163.

\bibitem{sparsegs}
H.~Xiong, S.~Muttukuru, H.~Xiao, R.~Upadhyay, P.~Chari, Y.~Zhao, and A.~Kadambi, ``Sparsegs: Sparse view synthesis using 3d gaussian splatting,'' in \emph{International Conference on 3D Vision 2025}, 2025.

\end{thebibliography}

\end{document}